\documentclass[conference]{IEEEtran}
\IEEEoverridecommandlockouts

\usepackage{cite}
\usepackage{amsmath,amssymb,amsfonts}
\usepackage{graphicx}
\usepackage{textcomp}
\def\BibTeX{{\rm B\kern-.05em{\sc i\kern-.025em b}\kern-.08em
    T\kern-.1667em\lower.7ex\hbox{E}\kern-.125emX}}

\usepackage{amsmath,amsfonts,bm}

\newcommand{\mbf}[1]{\mathbf{#1}}

\def\eqref#1{equation~\ref{#1}}

\def\1{\bm{1}}

\def\rvr{{\mathbf{r}}}
\def\rvs{{\mathbf{s}}}

\def\rvx{{\mathbf{x}}}

\def\va{{\bm{a}}}

\def\vv{{\bm{v}}}

\DeclareMathAlphabet{\mathsfit}{\encodingdefault}{\sfdefault}{m}{sl}
\SetMathAlphabet{\mathsfit}{bold}{\encodingdefault}{\sfdefault}{bx}{n}

\def\gC{{\mathcal{C}}}

\def\gE{{\mathcal{E}}}

\def\gL{{\mathcal{L}}}
\def\gM{{\mathcal{M}}}

\def\gR{{\mathcal{R}}}

\def\gT{{\mathcal{T}}}
\def\gU{{\mathcal{U}}}

\def\sR{{\mathbb{R}}}

\newcommand{\sethree}{\mathrm{SE(3)}}
\newcommand{\SE}{\mathrm{SE(3)}}

\newcommand{\sothree}{\mathrm{SO(3)}}
\newcommand{\SO}{\mathrm{SO(3)}}

\newcommand{\sothreelie}{\mathfrak{so}(3)}

\newcommand{\igso}{\mathcal{IG}_\mathrm{SO(3)}}

\newcommand\scalemath[2]{\scalebox{#1}{\mbox{\ensuremath{\displaystyle #2}}}}

\newcommand{\backbone}{\mathcal{R}}

\def\x{{\mathbf{x}}}
\def\y{{\mathbf{y}}}
\def\rmd{\mathrm{d}}

\usepackage{siunitx} 

\usepackage{etoc}
\etocdepthtag.toc{mtchapter}
\etocsettagdepth{mtchapter}{subsection}
\etocsettagdepth{mtappendix}{none}

\usepackage[resetlabels]{multibib}
\usepackage{multirow}
\usepackage{multicol}
\usepackage{enumitem}
\usepackage[normalem]{ulem}

\usepackage[dvipsnames]{xcolor}
\usepackage{threeparttable}
\usepackage{adjustbox}

\usepackage{framed} 
\definecolor{shadecolor}{rgb}{0.94, 0.97, 1.0}

\definecolor{cite_color}{HTML}{114083}
\definecolor{link_color}{RGB}{153, 0,0}  
\definecolor{url_color}{RGB}{153, 102,  0}
\definecolor{emp_color}{RGB}{0,0,255}
\definecolor{aliceblue}{rgb}{0.94, 0.97, 1.0}
\usepackage{hyperref} 
\hypersetup{
 colorlinks,
 citecolor=cite_color,
 linkcolor=link_color,
 urlcolor=url_color}
\usepackage{amsmath} 

\usepackage{caption}

\usepackage{pifont}

\usepackage{makecell}
\usepackage{subfigure} 
\usepackage{amsmath}
\usepackage{multirow}
\usepackage{multicol}
\usepackage{enumitem}
\usepackage{ulem}
\usepackage{threeparttable}
\usepackage{wrapfig}
\usepackage{scrextend}
\usepackage{thm-restate}
\usepackage{array,hhline}
\usepackage{amsmath,amsfonts}
\usepackage{tabularborder}
\usepackage{algorithm}
\usepackage{setspace}
\usepackage{mathtools}

\usepackage[commentColor=black,beginLComment=/*~, endLComment=~*/]{algpseudocodex}

\usepackage{xspace}
\newcommand{\flowab}{\text{FlowAB}\xspace}
\newcommand{\FlowAB}{\text{FlowAB}\xspace}

\definecolor{myred}{RGB}{215,48,39}
\definecolor{mygreen}{RGB}{26,152,80}
\newcommand{\cmark}{\textcolor{mygreen}{\ding{51}}}

\newcommand{\up}[1]{ {\color{myred}\uparrow  {#1}}}

\newcommand{\citep}[1]{\cite{#1}}

\begin{document}

\title{Efficient Antibody Structure Refinement \\ Using Energy-Guided SE(3) Flow Matching\\
\thanks{This work was accepted as a regular paper by BIBM 2024.}
}

\makeatletter
\newcommand{\linebreakand}{%
  \end{@IEEEauthorhalign}
  \hfill\mbox{}\par
  \mbox{}\hfill\begin{@IEEEauthorhalign}
}
\makeatother

\author{
Jiying Zhang,~Zijing Liu$^*$\thanks{$^*$Corresponding author},~Shengyuan Bai,~He Cao,~Yu Li$^*$, Lei Zhang\\
 International Digital Economy Academy (IDEA) \\
  Shenzhen, China \\
  \texttt{\{zhangjiying,liuzijing,baishengyuan,caohe,liyu,leizhang\}@idea.edu.cn}

}

\maketitle

\begin{abstract}
Antibodies are proteins produced by the immune system that recognize and bind to specific antigens, and their 3D structures are crucial for understanding their binding mechanism and designing therapeutic interventions.
The specificity of antibody-antigen binding predominantly depends on the complementarity-determining regions (CDR) within antibodies.
Despite recent advancements in antibody structure prediction, the quality of predicted CDRs remains suboptimal.  
In this paper, we develop a novel antibody structure refinement method termed FlowAB based on energy-guided flow matching. FlowAB adopts the powerful deep generative method $\sethree$ flow matching and simultaneously incorporates important physical prior knowledge into the flow model to guide the generation process. 
The extensive experiments demonstrate that \FlowAB can significantly improve the antibody CDR structures. It achieves new state-of-the-art performance on the antibody structure prediction task when used in conjunction with an appropriate prior model while incurring only marginal computational overhead. This advantage makes FlowAB a practical tool in antibody engineering. 

\end{abstract}


\section{Introduction}
Antibodies play a vital role in the immune system~\cite{neumeier2022phenotypic} and have been extensively utilized as therapeutics~\cite{carter2018next,reddy2010monoclonal}. Molecular structure determines its function~\cite{jing2024alphafold,zhengyang2024controlmol,gennis2013biomembranes,zhang2024subgdiff}. Understanding their structure is crucial to comprehending their antigen-binding ability. The six hypervariable complementarity determining regions (CDRs) are the primary sites for antigen binding and exhibit significant structural variation~(cf. \autoref{fig:cdr_loopers}), with the CDR-H3 and CDR-L3 loops being the most variable and critical region for antigen binding. Accurately modeling the structural variability of CDRs from sequence poses a significant challenge. Fortunately, recent advancements in deep learning methods have revolutionized our capacity to predict CDR structures.

\begin{figure}[th]
    \centering
    \includegraphics[width=0.9\linewidth]{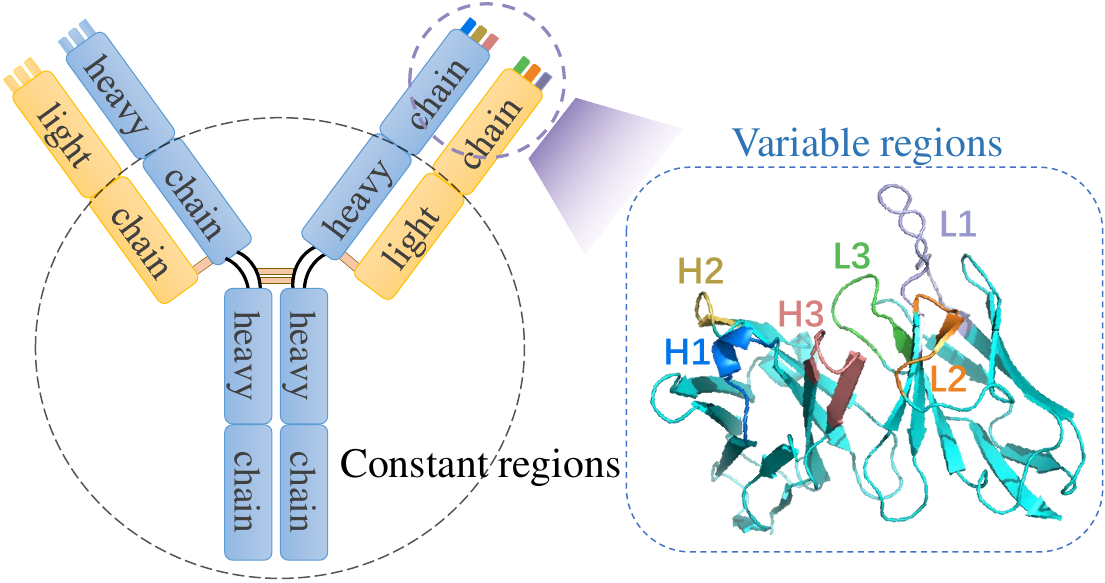}

    \caption{Illustration of the six Complementarity-Determining Regions (CDRs) of an Antibody. The regions labeled are CDR-L1, L2, L3 on the light chain, and CDR-H1, H2, H3 on the heavy chain. These CDRs, being highly variable and playing a critical role in antigen recognition, are the focus of structural refinement in this work.}  \label{fig:cdr_loopers}
     
\end{figure}

Current antibody structure prediction methods, such as IgFold~\cite{ruffolo2023fast}, have focused on \textit{de} \textit{novo} prediction of the antibody backbone structure, without utilizing the existing CDR structures.
These approaches leave room for improving the accuracy of diverse CDR structures, particularly the CDR-H3. Therefore, refining the CDR structures predicted by efficient, prior state-of-the-art methods like ABlooper~\cite{abanades2022ablooper} is promising for their application in biomedical research.

Conventionally, protein structure refinement involves several typical methods, including molecular dynamics (MD) simulation, energy minimization, and fragment assembly. MD-based methods~\cite{heo2021physics,lee2019galaxyrefine2} have demonstrated success by adopting physics-based approaches to sample multiple MD trajectories following the fundamental principles of atomic interactions. 

Energy minimization-based techniques~\cite{bhattacharya20133drefine} focus on optimizing protein structures by repacking both backbone and side-chain atoms, utilizing composite physics and knowledge-based force fields. 
Fragment assembly-based methods, \textit{e.g.} Rosetta~\cite{hiranuma2021improved}, which leverage template fragment information from the Protein Data Bank (PDB) and statistical potentials, share similarities with knowledge-based approaches.

Despite their effectiveness in refining specific protein structures with the help of comprehensive psychical information, these methods require significant computational resources and extensive conformation sampling.
Recently, deep learning has been applied to protein structure refinement with notable efficiency~\cite{jing2021fast,wu2022atomic}. However, those methods are designed for general protein structures and their performance of antibodies has not been thoroughly tested. Additionally, these methods cannot incorporate essential physical prior information, such as the inter-atomic potentials. This limits their ability to generate reasonable antibody structures that adhere to the physical principles. Given these challenges, a natural question arises: \textit{How can we design a deep learning method that incorporates physical prior information to efficiently optimize existing antibody structures}?
 
Flow model is a popular generative model known for its impressive sampling efficiency~\cite{chen2018neural,lipman2023flow} and has been successfully applied in various domains~\cite{bose2024sestochastic,stark2023harmonic}.
It assumes a deterministic continuous-time generative process and fits an ordinary differential equation (ODE) that transforms the source density into the target density. It is a promising method to refine the antibody structure by transforming the initial antibody structure to the optimized one.
In this paper, we design a new energy-guided flow model \textbf{\FlowAB} for antibody refinement (optimization).
\FlowAB consists of two important components: $\sethree$ flow matching~\cite{bose2024sestochastic} and energy guidance. Specifically, flow matching provides an efficient simulation-free manner for training \FlowAB through regressing the vector field, reducing the complexity of learning the distribution transformations. Simultaneously, the energy functions related to the physical prior information are incorporated into the flow to guide the generative process, enabling the generation of high-fidelity antibody CDRs that better match the native structure. 

Our experiments demonstrate that the proposed methods can efficiently optimize the antibody structure and also achieve state-of-the-art (SOTA) performance on the antibody structure prediction task.

The main contributions of our work are summarized as follows:
\begin{itemize}[leftmargin=5mm,noitemsep]
    \item We propose {\FlowAB}, a novel flow-based generative model specifically designed for antibody structure refinement, making the first successful attempt in this field.
    \item Energy guidance is integrated in {\FlowAB} to leverage physical prior information during the generative process, enabling more precise optimization of antibody structures.
    \item  The extensive experiments on antibody structure prediction and refinement demonstrate the superior performance of our model. With an appropriate prior and minimal additional computation, our method achieves SOTA performance on the SAbDab dataset.   
\end{itemize}

\section{Related Work}
\subsection{Flow matching generative model}

Recently, a new paradigm of generative models called flow matching has been introduced~\cite{albergo2022building,lipman2023flow}. It is a simplified framework extended from the diffusion model~\cite{ho2020denoising,song2020score} that offers more flexibility in design.
Recent works~\cite{esser2024scaling,pooladian2023multisample} demonstrate its ability to learn the flow between arbitrary source and target distributions in a simulation-free way. 
Meanwhile, Riemannian flow matching is proposed to handle the data on manifolds~\cite{chen2024flow}, which makes flow matching applicable in many biomolecular problems such as
protein structure generation~\cite{bose2024sestochastic,jing2024alphafold,yim2023se}, protein-ligand binding structures~\cite{stark2023harmonic}, and small molecule generation~\cite{song2023equivariant}.
The unique ability of flow matching to transform between two distributions presents a significant opportunity for refining existing protein structures. This is particularly relevant for antibodies, which feature highly diverse CDRs. Given this potential, there is a compelling case for developing a flow-matching-based model specifically tailored for optimizing antibody structures.

\subsection{Antibody structure prediction with deep learning}
Deep learning has revolutionized protein structure prediction, including the specialized area of antibody structures. For instance,
DeepAb~\cite{ruffolo2022antibody} utilizes interresidue geometric constraints to generate a comprehensive FV (variable fragment) structure, which is subsequently refined using Rosetta. ABlooper~\cite{abanades2022ablooper} employs an end-to-end approach to predict CDR loop structures, supplemented by post-prediction refinement. DiffAB~\citep{luo2022antigen} utilizes the $\sethree$ diffusion model to design and predict the CDR loop structure.
IgFold~\cite{ruffolo2023fast} directly predicts the coordinates of the backbone atom by an invariant point attention network and a pre-trained language model trained on antibody sequences. 
Despite these advancements, existing methods primarily focus on \textit{de} \textit{novo} structure prediction rather than refining suboptimal structures.
Further, compared with the diffusion-based model that requires extensive computational resources and demands thousands of inference steps, our \FlowAB only needs very few sampling iterations, thereby largely increasing the efficiency of antibody structure prediction.

\section{Preliminaries}

\subsection{Antibody backbone parameterization}

\begin{figure}[H]
    \centering
            \vspace{-4mm}
    \includegraphics[width=0.8\linewidth]{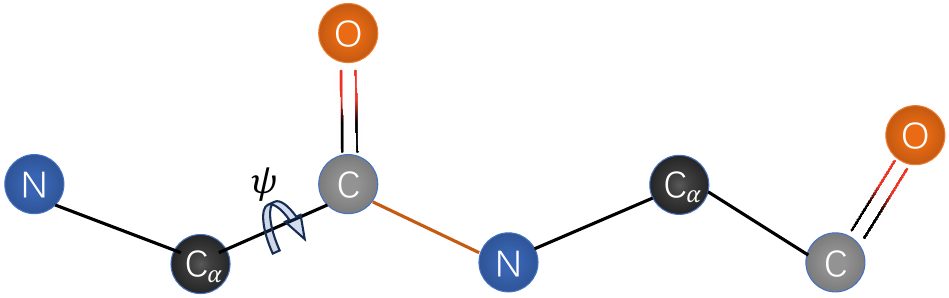}
    \caption{Illustration of the parametrization of the antibody backbone.}    \label{fig:protein_structure}
    \vskip - 0.1 in
\end{figure}
Our protein backbone parameterization draws inspiration from the seminal work of AlphaFold2~\citep{jumper2021highly}  and DiffAB~\citep{luo2022antigen}. We associate each residue in the amino acid sequence with a frame, resulting in $N$ $\sethree$-equivariant frames for a protein of length $N$. Denote the atom coordinates of $\text{C}_\alpha$ in residue $i\in [N]$  as $\rvx_i \in \mathbb{R}^3$, and its orientation as $\rvr_i \in \operatorname{SO(3)}$. Thus, 
residue $i\in [N]$ can be represented as an action of $\gR^i = (\rvx^i,\rvr^i) \in \sethree$ applied to the idealized frame $[\text{N},\text{C}_{\alpha},\text{C}]^i = \gR^i \circ [\text{N}^*,\text{C}^*_{\alpha},\text{C}^*]$. To construct the backbone oxygen atom $\text{O}$, we utilize $\text{N}^{i+1}$ rotation about the axis given by the bond between $\text{C}_{\alpha}$ and $\text{N}^{i+1}$ using an additional rotation angle $\psi$, which can be obtained from the Dhieral angle between $[\text{N}, \text{C}_\alpha]$ and $[\text{C},\text{N}]$. Finally, we denote the full 3D coordinates of the three heavy atoms in the frame ($[\text{N},\text{C}_{\alpha},\text{C}]$) as $\mbf{F} \in \sR^{N \times 3 \times 3}$. An illustration of this backbone parameterization is provided in \autoref{fig:protein_structure}.

\subsection{Problem definition of antibody optimization}
In this study, we focus on optimizing the structure of CDRs on an antibody framework, based on the given antigen and antibody structure. We assume that the generated CDR is a sequence of $m$ amino acids, with indices ranging from $l+1$ to $l+m$, denoting as $\gR=\{(\x_j,\rvr_j)\}_{j=k+1}^{k+m}$.
The structure and sequence of the antibody-antigen complex are denoted as $\gC = \{(\rvx_i, \rvr_i) \mid i \in \{1 \cdots N\} \setminus \{k + 1,\cdots, k + m\} \} + \{\rvs_i\}_{i=1}^{N}$, where $\rvs_i$ denotes the type of residue $i$.
The setting of antibody refinement is: Given an initial antibody structure $\gR_0$ generated by an existing model such as ABlooper~\cite{abanades2022ablooper}, the desired model $f$ will optimize $\gR_0$ to obtain an improved structure $\gR_1 =f(\gR_0, \gC)$ closer to the native structure. Due to the most diverse part of the antibodies being CDR regions, we focus on the optimization of the six CDRs.

\subsection{ Flow matching on Riemannian manifolds}
Flow matching (FM) is a cutting-edge simulation-free approach for training continuous normalizing flows (CNFs), a powerful class of deep generative models that integrate an ordinary differential equation (ODE) over a learned vector field. 
Recent advances have extended FM to handle Riemannian manifolds~\cite{chen2024flow}.
In particular, on a manifold $\gM$, the CNF $\phi_t(\cdot): \gM \to \gM$ is defined as an ODE associated with a time-dependent vector field $v(z,t): \gM\times [0,1] \to \gT_z\gM$, where $\gT_z\gM$ is the tangent space of the manifold at $z\in \gM$:
\begin{align}
\label{eq:CNF}
    \frac{d}{dt}\phi_t(z_0) = v_t(\phi_t(z_0),t); \quad \phi_0(z_0) = z_0.
\end{align}
When $z_0$ is sampled from a simple prior distribution $p_0$, \autoref{eq:CNF} can induce a probability path $p_t$ through the push-forward equation: $p_t=[p_t]*p_0$. The goal is to get the vector field $v$ that can generate the target data distribution $p_1$. However, a closed-form solution for $v$ is generally not available. Thus, \cite{lipman2023flow} proposes to learn $v$ by regressing the conditional vector field $u(\phi_t(z_0|z_1)| z_0,z_1) = \frac{d}{dt} \phi_t(z_0|z_1)$, where $\phi_t(z_0|z_1):=z_t$ interpolates between $z_0\sim p_0$ and $z_1\sim p_1$. On a simple manifold, $z_t$ can be calculated through the geodesic path $z_t = \exp_{z_0}(t\log_{z_0}(z_1))$. Here, $\exp_{z_0}$ and $\log_{z_0}$ are the exponential and logarithmic maps at the point $z_0$~\cite{chen2024flow}. The induced vector filed is: $u_t(z_t|z_1)=\log_{z_t}(z_1)/(1-t)$. Assume that $t\sim \gU([0,1])$ and $|\cdot|_g$ represent the norm induced by the Riemannian metric $g: \gT \gM \times \gT \gM \to \sR$. With the fact that learning the unconditional vector field $v$ is equivalent to learning the conditional vector field $u$, one can use a neural network ${v}_\theta$ to regress $u$:
\begin{align}
\label{eq:cfm}
    \gL = {\mathbb{E}}_{p_1(z_1),p_t(z_t|z_1)} \|{v}_\theta(z_t,t) -  u_t(z_t|z_1) \|_g^2.
\end{align}
Once the vector field $v_\theta$ is learned, samples can be generated by solving the ODE in \autoref{eq:CNF} with ${v}_\theta$ substituted for $v$.

\section{FlowAB: An Energy-Guided Flow Matching for Antibody Refinement}
\subsection{ SE(3) flow matching for backbone generation}
The $\sethree$ flow matching, a Riemannian flow matching, is developed in~\cite{chen2024flow} and   ~\cite{bose2024sestochastic}. Here, we use the $\sethree$ flow matching framework for generating the backbone of the antibody. The $\sethree$ can be decomposed into $\sothree$ and $\sR^3$ and then the vector field becomes $v_{\sethree} = (v_{\sR^3}, v_{\sothree})$. These components are utilized for modeling the $\text{C}_\alpha$ positions and amino acid (residue) orientations as follows.
This allows us to accurately capture both the spatial positions and the rotational alignments of the residues, crucial for correctly predicting the three-dimensional structure of the antibody.

\paragraph{Flow matching for $\text{C}_\alpha$ position.} Given the initial atomic coordinates $\x_0$ from the antibody structure prior distribution $p_0$ and the ground truth coordinates $\x_1 \sim p_1$, the probability path defined by interpolating linearly between $\x_0$ and $\x_1$ is $\x_t = (1-t)\x_0 + t\x_1 $. Then 
the probability path is associated with the vector field:
\begin{align}
    u_t(\x|\x_1) = \frac{\x_1-\x_t}{1-t} = \x_1- \x_0.
\end{align}
Thus, the training objective is:  
\begin{align}
\label{eq:R3_loss}
    \gL_{\sR} = \mathbb{E}_{p_1(\x_1),p_t(\x_t|\x_1)}\|v_\theta(\x_t,t,\gC) -u_t(\x|\x_1)\|_2^2.
\end{align}
\paragraph{$\sothree$ flow matching for residue orientations.}
 Given orientation $\rvr_0, \rvr_1 \in \sothree$ from a given initial sub-optimal antibody structure and native antibody structure, the probability path is defined as the geodesic interpolant form $\rvr_t = \exp_{\rvr_0}(t\log_{\rvr_0}(\rvr_1))$~\cite{chen2024flow}
and the corresponding vector field is: 
\begin{align}
    u_t(\rvr|\rvr_1) = \frac{\log_{\rvr_t}{\rvr_1}}{1-t}.
\end{align}
Thus, the training objective is:  
\begin{align}
\label{eq:SO3_loss}
    \gL_{\sothree} = \mathbb{E}_{p_1(\rvr_1),p_t(\rvr_t|\rvr_1)}\|v_\theta(\rvr_t,t,\gC) -u_t(\rvr|\rvr_1)\|_{\sothree}^2.
\end{align}

In this work, we do not employ the optimal transport plan $\pi$ calculated from the Monge and Kantorivich problem $\text{OT}(p_0,p_1)$ as the joint distribution to sample $p_0(\rvr_0), p_1(\rvr_1)$ for training~\cite{bose2024sestochastic}, because we assume that the residue $i$ in initial antibody structure given by a SOTA method has a good alignment with the residue $i$ in the target antibody structure.
The rationale for this decision is based on the assumption that the residue $i$ in the initial antibody structure, provided by a SOTA method, already exhibits a good alignment with the corresponding residue in the target antibody structure. This allows us to bypass computationally intensive optimal transport calculations, while maintaining alignment accuracy between the corresponding residues in the initial and target structures.

\begin{figure}[!t]
    \centering

    \includegraphics[width=1\linewidth]{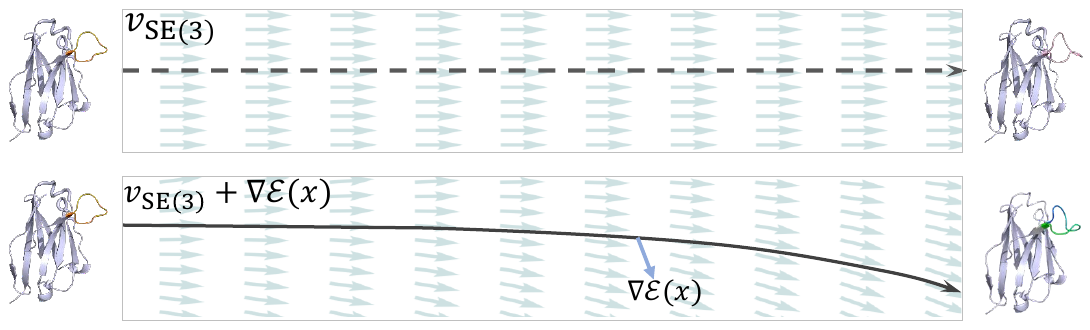}
     \caption{
     Flow matching with an energy guidance term $\nabla \mathcal{E}(x)$. The energy function gives guidance for the flow to reach a more physically plausible sample.
     } 
    \label{fig:EG_Trajectory}
\end{figure}

\subsection{Energy-guided flow matching}
For antibody structure refinement, initial structures produced by SOTA CDR predictors present a challenge due to the minor differences between them and native structures. This subtlety complicates the task of learning an effective vector field for structural adjustments.
Existing flow models, however, do not consider utilizing crucial physical prior information, such as the interatomic potential, to guide their sampling process. This omission limits their ability to robustly optimize the antibody structure when the vector field alone provides insufficient guidance.

To address these challenges, we introduce a novel energy-guided flow model designed to generate high-fidelity antibody conformations that can better match the molecular physical property. We propose incorporating human preferences to modulate the desired distribution and generate samples through ODE sampling with additional guidance. Inspired by the energy-guided technique in diffusion models~\cite{lu2023contrastive,wang2024protein}, we make the intermediate state $\backbone_t$ sample from the desired distribution:
\begin{align}
\label{eq:energy_prob}
q_t(\backbone_t) = p_t(\backbone_t)\frac{e^{-\beta\gE_t(\backbone_t)}}{Z},
\end{align}
where $Z:= \int p_t(\backbone_t)\exp(-\beta\gE(\backbone_t))d\backbone_t$ is the intractable normalizing constant, $\gE_t(\backbone_t)$  is the intermediate energy function from $\sethree$ to $\mathbb R$. $\beta \ge 0$ represents the inverse temperature, which controls the energy strength. The high-density region of the probability distribution $q_t(\backbone_t)$ is approximately the intersection of the high-density regions of $p_t(\backbone_t)$ and $e^{-\beta \mathcal E(\backbone_t)}$. 

Next, we need to design a new flow to get the probability path that complies with the \autoref{eq:energy_prob}.
Inspired by~\cite{yim2024improved}, we connect flow matching to diffusion models and adopt the probability flow ODE~\citep{song2020score} as follows:
\begin{align} \label{eq:score_ode}
    \scalemath{0.94}{ \rmd \backbone_t
    = \Tilde{v}(\backbone_t, t) \rmd t
    = \left[f(\backbone_t, t) - \frac{1}{2} g(t)^2 \nabla \log p_t(\backbone_t) \right] \rmd t,}
\end{align}
where $f$ and $g$ are the drift and diffusion coefficients, respectively. By sampling from the prior distribution $p_0$ and subsequently integrating the learned vector field $\Tilde{v}$ that is trained through minimizing the objective in \autoref{eq:cfm}, the ODE in \autoref{eq:score_ode} can get samples from $p_t(\backbone_t)$. This is the probability flow ODE that shares the same marginal probability as the stochastic differential equation (SDE) of the diffusion 
 model~\citep{song2020score}. This vector filed contains score $\nabla \log p_t(\backbone_t)$, to which the energy guidance can
be applied.
In particular, modifying \autoref{eq:score_ode} to involve the energy function of \autoref{eq:energy_prob}:
\begin{align}
\scalemath{0.9}{
    \rmd \backbone_t}
    &\scalemath{0.9}{= \left[f(\backbone_t, t) - \frac{1}{2} g(t)^2 \nabla \log (p_t(\backbone_t)e^{-\beta\gE_t(\backbone_t)}) \right] \rmd t }\\
    &\scalemath{0.9}{= \left[f(\backbone_t, t) - \frac{1}{2} g(t)^2 \left( \nabla \log p_t(\backbone_t) - \nabla \beta\gE_t(\backbone_t) \right) \right] \rmd t \nonumber} \\ 
    &\scalemath{0.9}{= \Bigl[\underbrace{\Tilde{v}_{\SE}(\backbone_t, t)}_{\text{unconditional vector field}}  + \tfrac{1}{2} g(t)^2\underbrace{~ \nabla \beta\gE_t(\backbone_t)}_{\text{guidance term}} \Bigr] \rmd t,} \label{eq:energy_ode}
\end{align}
where $g(t)$ can be used as a scheduler that controls the 
magnitude of the guidance.
We can interpret \autoref{eq:energy_ode} as doing unconditional generation by following $\hat{{v}}_\SE(\backbone, t)$ while $\nabla \beta\gE_t(\backbone_t)$ guides the residues by giving the force to the velocity (cf. Appendix \ref{appsubsec:dynamic_system}).
Following $\SE$ flow matching, \autoref{eq:energy_ode} becomes the following:
\begin{align}
    \text{C}_\alpha ~\text{position: }  \rmd \x_t &= \Bigl[\Tilde{v}_{\sR^3}(\rvx_t, t) + \tfrac{1}{2} g(t)^2\nabla_{\x_t}\gE(\backbone_t) \Bigr] \rmd t,     \label{eq:r3_recon}\\
    \text{Orientation: } \rmd \rvr_t &= \Bigl[\Tilde{v}_{\SO}(\rvr_t, t) + \tfrac{1}{2} g(t)^2\nabla_{\rvr_t}\gE(\backbone_t) \Bigr] \rmd t.   \label{eq:so3_recon}
\end{align}
In practice, we still utilize the training objectives presented in \autoref{eq:R3_loss} and \autoref{eq:SO3_loss} to learn the unconditional vector fields $\Tilde{v}_{\sR^3}(\rvx_t, t)$ and $\Tilde{v}_{\SO}(\rvr_t, t)$. Once the vector field is learned, there is no need for further model retraining. Instead, we can directly incorporate the energy-guided term during the sampling process. This approach becomes particularly valuable when an unconditional generative flow model is readily available, and further training is prohibitively expensive. In practice, the $\rmd t$ will be changed into $-\rmd t$ due to the time $t$ from $0$ to $1$ in sampling. Further explanations can be found in Appendix \ref{appsub:implement}.

\subsection{Interatomic potential as energy}
\begin{figure}[!h]
    \centering
    \includegraphics[width=0.64\linewidth]{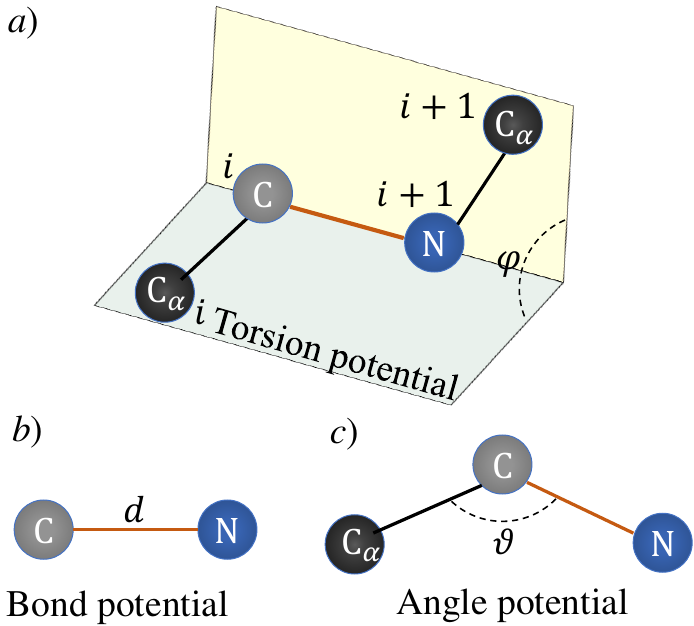}
     \caption{Illustration of inter-atomic potentials: \textit{a}) the improper torsion potential;
     \textit{b}) the bond potential; \textit{c}) the angle potential.}
    \label{fig:potential_energy}
\end{figure}
\begin{figure*}[ht]
    \centering

    \includegraphics[width=0.85\linewidth]{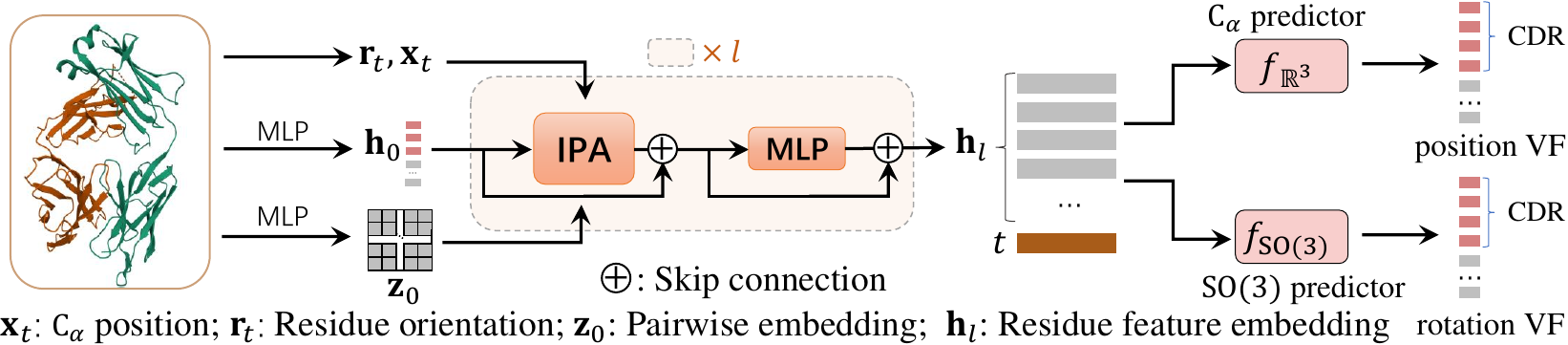}
     \caption{{ 
     The model architecture of \FlowAB. The model takes in the residue embedding $\mathbf{h}_0$, pairwise embedding $\mathbf{z}_0$, SO(3) vector $\mathbf{r}_t$ and $\mathbb{R}^3$ vector $\mathbf{x}_t$ at time $t$. 
     } }
    \label{fig:model_architecture}
     
\end{figure*}
From the antibody refinement perspective, the $\sethree$ flow on each residue depends only on its vector field, with $\sR^3$ flow refining the global position and $\sothree$ flow adjusting the local frame. However, these flows focus on individual residues and do not account for interactions with adjacent residues, potentially compromising the protein's physical properties in 3D space. Therefore, an energy-guided term is crucial for constraining the flow to produce physically plausible models. Motivated by molecular dynamics (MD), we consider bond, angle, and torsion potentials as physics-based preference functions.

\paragraph{Bond potential}
The bond potential (\autoref{fig:potential_energy} $b$) describes an oscillation about an equilibrium bond length $d_0$ with bond constant $k_b$: 
\begin{align}
    \gE(d)=k_b(d_{ij} - d_0)^2,
\end{align}
where $d$ is the distance between the atoms, 
$d_{0}$ is the equilibrium bond distance. 

\paragraph{Angle potential}
This potential~(\autoref{fig:potential_energy} $c$) describes oscillation about an equilibrium angle $\vartheta_0$ with force constant $k_{\vartheta}$:
\begin{align}
    \gE(\vartheta)=k_\vartheta(\vartheta-\vartheta_0)^2.
\end{align}
When the lengths of the two bonds that form the angle are known, the angle potential can be implemented by a three-bond potential among the three atom pairs. For example, the angle potential on $\vartheta$ can be equivalent to bond potential in $\text{C}-\text{C}_\alpha$, $\text{C}-\text{N}$ and virtual bond $\text{C}_\alpha -\text{N}$.

\paragraph{Improper torsion potential}
This potential is usually used to enforce planarity. It can be defined as a harmonic function:
\begin{align}
    \gE(\varphi ) = k_\varphi(\varphi-\varphi_0)^2,
\end{align}
where $\varphi$ is the dihedral angle between two planes. As shown in $a$) of \autoref{fig:potential_energy}, the dihedral angle $\varphi$ is the angle between planes $[\text{C}_\alpha^i,\text{C}^i, \text{N}^{i+1}]$ and  $[\text{C}^i, \text{N}^{i+1},\text{C}_\alpha^{i+1}]$.

\paragraph{Constructing potential function for antibody refinement}
\begin{figure}[H]
    \centering
    \includegraphics[width=0.95\linewidth]{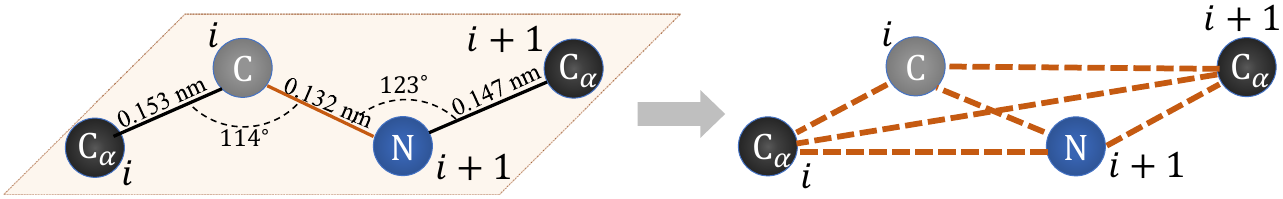}
    \caption{Constraining four atoms $[\text{C}_\alpha^i,\text{C}^i,\text{N}^{i+1},\text{C}_\alpha^{i+1}]$ between residue $i$ and $i+1$ on the amide plane would be equivalent to endow six bond potentials into each atomic pairs. }    \label{fig:six_bonds}
\end{figure}

Amino acids are connected through peptide bonds, which are amide linkages between the $-\text{NH}_2$ and $-\text{COOH}$ groups of adjacent amino acids. The peptide bond $\text{C}-\text{N}$ exhibits partial double bond characteristics, making it rigid and planar, and therefore not free to rotate. The plane in which the peptide bond resides is referred to as the peptide plane or amide plane~\cite{choudhuri2014bioinformatics}. Here, we aim to add the energy to guide the residues so that they have reasonable linkages. Specifically, given the residue $i$ and $i+1$, the following potentials would be considered:
\begin{itemize}[leftmargin=14pt, noitemsep]
    \item  A bond potential on the peptide bond $\text{C}^i-\text{N}^{i+1}$, with $d_0 = 1.32$ \char"C5~\cite{ha2011essentials}. This potential can effectively avoid the break-chain problem.
    \item
    Two angle potentials in the angles formed by [$\text{C}^i_\alpha-\text{C}^i$, $\text{C}^i-\text{N}^{i+1}$] and [$\text{C}^i-\text{N}^{i+1}$, $\text{N}^{i+1}-\text{C}_\alpha^{i+1}$], with $\vartheta_0=114^{\circ}$ and $\vartheta_0=123^{\circ}$~\cite{ha2011essentials} respectively.
    \item
     An improper torsion on the four atoms $[\text{C}^i_\alpha,\text{C}^i, \text{N}^{i+1},\text{C}_\alpha^{i+1}]$ with $\varphi_0=180^{\circ} $ to make them to be located on the amide plane~\cite{choudhuri2014bioinformatics}.
\end{itemize}
In practice, since the length of bond $\text{C}^i_\alpha-\text{C}^i$ and bond $\text{N}^{i+1}-\text{C}_\alpha^{i+1}$ is known, these above potentials are equivalent to six bond potentials between those four atoms, as shown in \autoref{fig:six_bonds}. Furthermore, considering that the lengths of bond $\text{C}^i_\alpha-\text{C}^i$ and $\text{N}^{i+1}-\text{C}_\alpha^{i+1}$ are unrelated to $\x_i,\rvr_i$, we only need to calculate the rest four potentials for guiding the flow. The details of the energy functions can be found in Appendix~\ref{appsec:Details_EG}.

The energy terms introduced will provide the necessary dependency between the joint residues, ensuring more reasonable positioning in the refined structure. This will also enhance the robustness of \FlowAB, allowing it to handle errors better in the learned vector field.

\subsection{Model architecture}

Our main model architecture integrates the Invariant Point Attention (IPA)~\cite{jumper2021highly} for geometric feature extraction and Multiple Layer Perceptrons (MLPs) for feature embedding and vector field prediction~\cite{luo2022antigen}. MLPs generate embeddings for both individual amino acids and pairs of amino acids.
The MLP of a single amino acid creates an embedding vector $\mbf h_0$ for each amino acid, capturing critical data that include amino acid types, torsional angles, and 3D coordinates of heavy atoms. For pairwise embeddings, another MLP processes the Euclidean distances and dihedral angles between amino acids $i$ and $j$, producing vectors $\mathbf{z}^{ij}_0$. These embeddings are transformed to hidden representations $\mathbf{h}_l$ via the IPA modules~\cite{jumper2021highly}. Subsequently, two separate MLPs predict the vector fields for the positions and orientations of the CDRs. The framework is shown in \autoref{fig:model_architecture}.

\paragraph{Full Loss} Inspired by~\cite{yim2023se} and~\cite{bose2024sestochastic}, we incorporate an auxiliary loss $\gL_{2d}$ to obtain a good structure in Euclidean space. Specifically, our training objective is defined as follows:
\begin{align}
    \gL = \gL_{\sR} + \gL_{\sothree}  + \lambda\mbf{1}\{t>0.5\}\gL_{2d},
\end{align}
where $\gL_{2d} =\| \mbf{1}\{ \mbf D<0.6nm \}(\mbf D-\hat{\mbf D} )\|_2/(\sum \mbf{1}(\mbf{D}<0.6nm)-N)$  and $\mbf D\in \sR^{N\times N\times 3\times3}$ is the distance between the three heavy atoms [N,$\text{C}_\alpha$,C]: $\mbf D_{ijkl} = \mbf F_{ik}-\mbf F_{jl}$, and a similar calculation yield $\hat{\mbf D}$. $\mbf{1}$ here represents the indicator function and $\lambda$ is a trade-off coefficient. We also consider a simplified version that only calculates the $\text{C}_\alpha$ position in adjacent amino acids $\mbf D_{i,i+1} = \mbf x_{i}-\mbf x_{i+1}$.  The training process is summarized in Algorithm~\ref{alg:flowab_training}.

\vspace{-2mm}
\begin{algorithm}
\caption{\flowab training}
\label{alg:flowab_training}
\begin{algorithmic}[1]
\State \textbf{Input:} Source and target $p_0(\x_0), p_1(\x_1)$, flow network $v_\theta$
\While{Training}
    \State $t, \x_0, \x_1 \sim \gU(0, 1), p_0, p_1$
    
    \State $\rvr_t \leftarrow \exp_{\rvr_0}(t \log_{\rvr_0}(\rvr_1))$ \Comment{geodesic interpolant from
    }
    \State $\x_t \leftarrow t \x_0 + (1 - t) \x_1$
    \Comment{interpolant (Euclidean)}
    
    \State ${\mathcal{L}}_{\flowab} \leftarrow \left \|v_\theta( {\rvr}_t, t, \gC) - \frac{\log_{{\rvr}_t}(\rvr_1)}{1-t} \right \|_{\sothree}^2 + 
    \left \| v_\theta({\x}_t,t, \gC) - \frac{\x_1-{\x}_t }{1-t} \right\|^2 $
    \State $\theta \leftarrow \text{Update}(\theta, \nabla_\theta \mathcal{L}_{\flowab,})$
\EndWhile
\State \textbf{return} $v_\theta$
\end{algorithmic}

\end{algorithm}
\vspace{-5mm}
\begin{algorithm}
\caption{\FlowAB inference}
\label{alg:flowab_inference}
\begin{algorithmic}[1]
\State \textbf{Input:} Source distribution \( p_0 \), flow network \( v_\theta \), inference annealing \( i(\cdot) \), integration step size \( \Delta t \), energy guided weight $g(t)$.
\State Sample \( (\x_0,\rvr_0) \sim p_0 \)
\For{ \( s\) in \( (0, 1 / \Delta t] \)}
    \State $t \gets s \Delta t$
    \State \( u_t \leftarrow \rvr_t^\top (v_\theta( \rvr_t, t,\gC)- \frac{1}{2} g(t)^2\nabla_{\rvr_t}\gE(\backbone_t)) \)
    \\ \Comment {parallel-transport the vector field to $\sothreelie$}
    \State \( \rvr_{t+\Delta t} \leftarrow \rvr_t \exp(u_t i_t \Delta t ) \)\Comment {rotation update}
    \State  $\x_{t+\Delta t} \leftarrow  \x_t +( v_\theta(\x_t, t,\gC)- \frac{1}{2} g(t)^2\nabla_{\x_t}\gE(\backbone_t))\Delta t$ 
    
    \Comment {position update}
\EndFor
\State \textbf{return} \( \rvr_1, \x_1 \)
\end{algorithmic}
\end{algorithm}

\section{Experiments}
\subsection{Antibody structure refinement}
\paragraph{Settings and dataset} 
In line with standard antibody structure prediction settings, we utilize the widely used Structural Antibody Database (SAbDab)~\cite{dunbar2014sabdab}. To train the \FlowAB, the structures available in SAbDab were considered as the ground truth (target), and two types of prior samples (source) are generated by ABlooper~\cite{abanades2022ablooper} or DiffAB~\cite{luo2022antigen} respectively\footnote{Both methods are efficient (cf. Tab. \ref{tab:runtimes})  and can only predict the structure of CDRs. Therefore, they are suitable as priors in our experiments.}.
To ensure a fair comparison with different baselines, we performed data splits using two different versions: 1) The IgFold split~\cite{ruffolo2023fast}, which is a commonly used split for comparing state-of-the-art antibody prediction methods. This split consists of 197 test samples; 2) The DiffAB split~\cite{luo2022antigen}, which is specifically designed to predict the structure of CDRs only. The DiffAB split contains 19 test samples.
\begin{wrapfigure}{R}{0.6540\linewidth}
    \centering
  
    \includegraphics[width=0.99\linewidth]{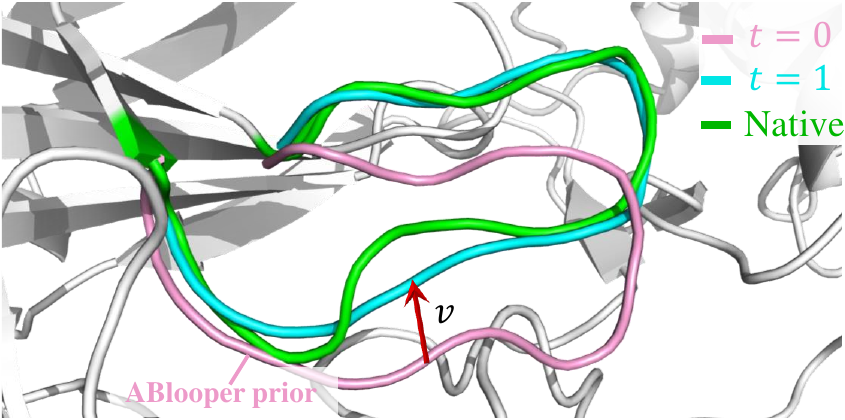}
     \caption{\FlowAB refines the structure of CDR-H3 from the ABlooper prior.}
    \label{fig:visulization_H3}
        \vspace{-1mm}
\end{wrapfigure}
The evaluation metric used in our experiments is the mean square deviation (RMSD) of the backbone structure. The detailed definition can be found in Appendix~\ref{appsubsec:rmsd}.
During training, we optimize only the CDRs while maintaining the rest of the regions as the initial structure. For sampling, we refine the structures of the CDRs in the test set. We set the number of time steps to be 2 (that is, $\Delta t=0.5$) and the whole sampling process is summarized in Algorithm~\ref{alg:flowab_inference}. More experimental details are available in Appendix~\ref{appsec:experiment_setup}.

\paragraph{Baselines}
We include several typical antibody structure prediction methods in our evaluation, namely RepertoireBuilder~\cite{ruffolo2023fast}, AlphaFold-Multimer~\cite{evans2021protein}, AlphaFold3~\cite{abramson2024accurate}, DeepAb~\cite{ruffolo2022antibody}, ABlooper~\cite{abanades2022ablooper}, DiffAB~\citep{luo2022antigen}, and IgFold~\cite{ruffolo2023fast}. Among these, ABlooper and DiffAB specifically focus on predicting the structure of CDRs. Additionally, we compared our approach with the MD-based refinement method Rosetta~\cite{alford2017rosetta}, implemented using Python.

\paragraph{Results}
From \autoref{tab:diffab_split} and \autoref{tab:igfold_split}, we have the following observations: 1) Our method consistently optimizes the structures produced from ABlooper in both data splits, suggesting that our methods can take advantage of the information in ABlooper to getting the more native structure (cf. \autoref{fig:visulization_H3}).
2) Combining with the efficient SOTA method ABlooper, our method can be viewed as a \textit{de novo} structure prediction framework. The framework is computationally efficient (cf. \autoref{tab:runtimes}) and achieves superior performance against the SOTA antibody prediction methods. 3) The results in \autoref{tab:diffab_split} indicate that \FlowAB can efficiently refine the initial structures from DiffAB and ABlooper, demonstrating the robustness and generalization of the proposed method.

\subsection{Ablation study}
\begin{wraptable}{r}{0.580\linewidth}
\vspace{-3mm}
\caption{\small{ 
The average runtimes for generating one structure (in $seconds$).
}
}
\vspace{-2mm}
\label{tab:runtimes}
\centering
\begin{adjustbox}{max width=\textwidth}
\small
\setlength{\tabcolsep}{8pt}
\centering
\begin{threeparttable}[b]
\begin{tabular}{l c c c c c c}
\toprule

Method & Time\\
\midrule
DiffAB & $1.95$\\
ABlooper & $1.23$ \\
\midrule
IgFold~(w. Rosetta) & $23$ \\

ABlooper~(w. OpenMM)  & $174$ \\

\midrule

 DiffAB + FlowAB& $2.18$ \\
  ABlooper + FlowAB & ${1.48}$ 
  \\
\bottomrule
\end{tabular}

\end{threeparttable}
\end{adjustbox}

\vspace{-3mm}
\end{wraptable}

We conduct an ablation study to validate the effectiveness of two key components in our approach: the flow matching term $\Tilde{v}_{\sethree}$ and the energy guidance $\nabla \gE$. The results presented in \autoref{tab:ablation_study} demonstrate that both flow matching and energy guidance contribute to improving the structure individually. Moreover, these two components work collaboratively, resulting in a consistent boost in the performance of antibody refinement.
\begin{table*}[!t]
\vspace{-1mm}
\caption{{
Results for antibody structure optimization (DiffAB split~\cite{luo2022antigen}). 
We report the average RMSD (\AA) in the test set.
The relative improvement by FlowAB is reported in {\color{myred}{red}}.
Note that ABlooper + Pyrosetta takes an average of about 450 seconds per sample, while ABlooper + FlowAB takes only 1.5 seconds, with FlowAB costing just 0.25 seconds.
}
}
\vspace{-1mm}
\label{tab:diffab_split}
\centering
\begin{adjustbox}{max width=\textwidth}
\setlength{\tabcolsep}{22 pt}
\centering
\begin{tabular}{l l l l l l l}
\toprule
Method & H1 $\downarrow$ & H2 $\downarrow$ & H3 $\downarrow$ & L1 $\downarrow$ & L2 $\downarrow$ & L3 $\downarrow$ \\
\midrule

DiffAB~\cite{ruffolo2022antibody} & $0.75$ &	$0.87$ &	$3.25$ &	$1.25$ &	$1.09$ &	$1.14$ 
 \\

ABlooper~\cite{abanades2022ablooper} & $0.75$ &	$0.72$ &	$2.45$ &	$1.00$ &	$0.88$ &	$1.07$
  \\
ABlooper + Pyrosetta & 0.66 & 0.65 & 2.42 & 0.96 & 0.85 & 1.14\\

\midrule 
DiffAB + FlowAB & $0.72$ $_{\up{4\%}}$ & $0.84$ $_{\up{3\%}}$ & $2.70$ $_{\up{17\%}}$ & ${1.22}$ $_{\up{2\%}}$ & $1.07$ $_{\up{2\%}}$ & $1.03$ $_{\up{10\%}}$ \\
ABlooper + FlowAB &  $0.60$ $_{\up{20\%}}$&	$0.70$ $_{\up{3\%}}$	& {$2.20$} $_{\up{10\%}}$&	$0.98$ $_{\up{2\%}}$& $0.87$ $_{\up{1\%}}$&	$0.99$ $_{\up{7\%}}$
\\

\bottomrule
\end{tabular}
\end{adjustbox}
\end{table*}

\begin{table*}[t]
\caption{\small{
Comparison with SOTA methods on antibody structure prediction  (IgFold split \cite{ruffolo2023fast}). 
We report the average RMSD (\AA) in the test set.
The best and second best results are marked in \textbf{bold} and \underline{underlined}.
}}
\vspace{-1mm}
\label{tab:igfold_split}
\centering
\begin{adjustbox}{max width=\textwidth}
\small
\setlength{\tabcolsep}{23pt}
\centering
\begin{threeparttable}
\begin{tabular}{l c  c c c c c}
\toprule
Method  & H1 $\downarrow$ & H2 $\downarrow$ & H3 $\downarrow$ & L1 $\downarrow$ & L2 $\downarrow$ & L3 $\downarrow$ \\
\midrule
RepertoireBuilder~\cite{schritt2019repertoire} & $1.00$ & $0.90$ & $4.15$ &  $0.81$ & $0.57$ & $1.32$ \\
DeepAb~\cite{ruffolo2022antibody} (w. Rosetta)  & $0.86$ & \uline{$0.72$}  & $3.57$ & \uline{$0.75$} & $0.48$ & $1.16$ \\
ABlooper~\cite{abanades2022ablooper} (w. OpenMM~\cite{eastman2017openmm})  & $0.98$	& $0.83$ &  $3.54$	& $0.92$ &	$0.67$ &	$1.32$	\\
ABlooper~\cite{abanades2022ablooper}  & $0.89$ &	$0.76$ &	\uline{$3.15$} &	$0.83$ &	$0.48$ &	\uline{$1.15$} \\
DiffAB~\cite{luo2022antigen}  &  $1.08$ &	$0.94$	& $3.94$	& $1.03$	 & $0.62$	& $1.49$ 	 \\
AlphaFold3$^\star$~\cite{abramson2024accurate}
&3.24 & 2.81 & 3.62 & 2.82 & 1.90 & 2.72 \\
AlphaFold-Multimer~\cite{evans2021protein}  & $0.95$ & $0.74$ & $3.56$ & $0.84$ & $0.51$ &$1.59$ \\
IgFold~\cite{ruffolo2023fast} (w. Rosetta)   & \uline{$0.85$}&	$0.76$	&{$3.27$}&	$0.76$&	\uline{$0.46$}&	$1.30$ \\

\midrule

ABlooper + FlowAB (Ours)   & $\mbf{0.82}$	& $\mbf{0.70}$ &  $\mbf{2.98}$	& $\mbf{0.70}$ &	$\mbf{0.38}$ &	$\mbf{1.03}$	\\

\bottomrule
\end{tabular}
\begin{tablenotes}
    \scriptsize
     \item[$\star$] 
     Antibody-antigen sequence pairs are entered as complexes into the \href{https://alphafoldserver.com/}{online server} for prediction.
     The quality of CDR prediction may be affected if the antigen-antibody interface is not correctly predicted.
     \end{tablenotes}
\end{threeparttable}
\end{adjustbox}

\vspace{-1mm}
\end{table*}

\begin{table*}[ht!]
\caption{\small{
Results for ablation study (DiffAB split~\cite{luo2022antigen}). 
We report the average RMSD (\AA) in the test set.
}
}
\vspace{-1mm}
\label{tab:ablation_study}
\centering
\begin{adjustbox}{max width=\textwidth}
\setlength{\tabcolsep}{16pt}
\centering
\begin{tabular}{l | c c | c c c c c c}
\toprule
Prior &Flow matching & Energy guidance& H1 $\downarrow$ & H2 $\downarrow$ & H3 $\downarrow$ & L1 $\downarrow$ & L2 $\downarrow$ & L3 $\downarrow$ \\
\midrule
 \multirow{4}{*}{DiffAB}
 & -- & -- & $0.7594$ & $0.8704$ & $3.2461$ & $1.2543$ & $1.0859$ & $1.1378$ \\
 & -- &  \cmark  & $0.7464$ & $0.8572$ & $3.2492$ & $1.2446$ & $1.0788$ & $1.1383$  \\
  & \cmark & -- & $0.7302$ & $0.8480$ & $2.7108$  &  $1.2415$ &$1.0749$ & $1.0390$\\

  & \cmark & \cmark  & $\mbf{0.7179}$ & $\mbf{0.8358}$  &  $\mbf{2.7040}$& $\mbf{1.2163}$ & $\mbf{1.0688}$& $\mbf{1.0339}$\\
 
\midrule 
\multirow{4}{*}{ABlooper}
 & -- & -- & $0.7465$ & $0.7239$ & $2.4497$ & $1.0012$ & $0.8821$ & $1.1471$ \\
 &  -- & \cmark& $0.7422$ & $0.7225$ & $2.4484$ & $0.9991$ & $0.8803$ & $1.1443$ \\
 & \cmark & -- & $0.5998$ & $0.6934$ & $2.2176$ & $0.9839$ & $0.8693$& ${0.9859}$ \\

 & \cmark & \cmark  & $\mbf{0.5965}$ & $\mbf{0.6927}$
 & $\mbf{2.2072}$ & $\mbf{0.9767}$ & $\mbf{0.8683}$
 & $\mbf{0.9855}$  
  \\

\bottomrule
\end{tabular}
\end{adjustbox}
\vspace{-3mm}
\end{table*}

\subsection{Structure prediction time analysis}

We compare the structure prediction time with the baseline to demonstrate the impressive efficiency of our method. For our benchmark, all deep learning methods were executed on identical hardware (a 32-core CPU with a single RTX 3090 GPU), ensuring a direct comparison of their runtimes.  In our experiment, we find that \FlowAB takes around $0.25 s$  on average to refine an antibody (2 steps). From~\autoref{tab:runtimes}, we can see that the combination ABlooper + FlowAB achieves better performance than the structure prediction methods IgFold and DiffAB, implying that this method can be used for efficiently \textit{de nove} antibody structure generation.

\section{Conclusion and limitations}\label{sec:conclusion} 

In this paper, we have proposed \FlowAB, a novel flow-based generative model for the refinement of the antibody structure. By incorporating the flow matching training objective and integrating the energy function into the flow model, \FlowAB facilitates the generation of antibody structures in a simple and efficient manner. However, one limitation of this method is that the energy terms used in \FlowAB employ a relatively simple potential function. The design of a more comprehensive energy function could further enhance the quality of the generated antibody structures.
In terms of future work, \FlowAB can be extended to support \textit{de novo} antibody design by incorporating a discrete flow component for the generation of amino acid types. This would allow the model to not only refine existing antibody structures but also generate novel antibody sequences and structures from scratch.

\bibliographystyle{IEEEtran.bst}
\bibliography{0_main.bib}
\newpage
\onecolumn
\appendix

\etocdepthtag.toc{mtappendix}
\etocsettagdepth{mtchapter}{none}
\etocsettagdepth{mtappendix}{subsection}
\tableofcontents   


\section{Fusing energy function into flow matching}

\subsection{Implement of energy-guided flow matching}\label{appsub:implement}
In our sampling process, the time step $t$ is from $0$ to $1$. While the \autoref{eq:energy_ode} describes an ODE from target to source (a denoising process), where $dt$ is a negative infinitesimal negative timestep~\cite{song2020score}. We rewrite the $\rmd t$ to be $-\rmd t$, then the ODE become 
\begin{align}
    \scalemath{0.88}{
    \rmd \backbone_t = \Bigl[\underbrace{-\Tilde{v}_{\SE}(\backbone_t, t)}_{\text{unconditional vector field}}  - \tfrac{1}{2} g(t)^2\underbrace{~ \nabla \beta\gE_t(\backbone_t)}_{\text{guidance term}} \Bigr] \rmd t.} \label{eq:energy_neagtive_ode}
\end{align}
Recall the training objective in \autoref{eq:cfm}, we actually regress the $-\Tilde{v}_{\SE}(\backbone_t, t)$ during training process in Alg.~\ref{alg:flowab_training}. Therefore, 
\begin{align}
      u_t &= \rvr_t^\top (v_\theta( \rvr_t, t,\gC)- \frac{1}{2} 
      g(t)^2\nabla_{\rvr_t}\gE(\backbone_t))\\
      \rvr_{t+\Delta t} &= \rvr_t \exp(u_t i_t \Delta t) \\
     \x_{t+\Delta t} &=  \x_t +( v_\theta(\x_t, t,\gC)- \frac{1}{2} g(t)^2\nabla_{\x_t}\gE(\backbone_t))\Delta t
\end{align}

\subsection{Understanding the energy guided flow from Molecular dynamic simulation perspective}\label{appsubsec:dynamic_system}
The generative process of antibody structure can be viewed as the evolutionary simulation of the system in time. Thus, we need to solve Newtonian equations of motions in the dynamic system:
\begin{align}
    \gR^{t+\delta t} = \gR^t + \vv(t)\delta t + \frac{1}{2} \va(t)\delta t^2
\end{align}
where $\vv(t)$ is the velocity and $\va(t)$ is the acceleration. Therefore, our energy-guided in \autoref{eq:r3_recon} and \autoref{eq:so3_recon} is equivalent to the above dynamic system when we set $g(t)^2 = dt =\delta t$.

\section{The computational details of energy guidance }\label{appsec:Details_EG}
In the antibody backbone, the global coordinates $\y$ of atoms in the backbone are accessed by $\y = \rvr \y^* + \x$.

We first concentrate on the calculation the gradient of bond potential between $\text{C}_\alpha^i-\text{C}_\alpha^{i+1}$.
\paragraph{I) $\text{C}_\alpha^i-\text{C}_\alpha^{i+1}$ energy guidance.}
We know $d_e=3.8$\char"C5. Let $j=i+1$,
\begin{align}
    d_{ij} &= \|\x^i - \x^j\|_2\\
    \gE(d_{ij}) & = k_{\alpha}(d_{ij} - d_e)^2 \\
    \nabla_{\x^i} \gE(d_{ij})  &= 2k_{\alpha} (d_{ij}-d_e) \frac{\x^i-\x^j}{d_{ij}}\\
    \nabla_{\x^j} \gE(d_{ij})  &= -2k_{\alpha} (d_{ij}-d_e) \frac{\x^i-\x^j}{d_{ij}}\\
\end{align}
The energy function of the CDRs is
\begin{align}
\gE_{1} &= \sum_{i}\sum_{j=i+1} \gE(d_{ij});\\
\nabla_{\x^i} \gE_{1} &= \sum_{i}\sum_{j=i+1} \nabla_{\x^i}\gE(d_{ij});
\end{align}

\paragraph{II) The energy guidence between $\text{C}^i-\text{N}^{i+1}$. }
Here, $d_e=1.32$\char"C5. Let j=i+1
\begin{align}
\y^i &= \rvr^i \y^{i*} + \x^i; \frac{d\y^i}{d\x^i} = \mbf{1} \\
\y^j &= \rvr^j \y^{j*} + \x^j;\frac{d\y^j}{d\x^j} = \mbf{1} \\
    d_{ij} &= \|\y^i - \y^j\|_2\\
    \gE(d_{ij}) & = k_{\alpha}(d_{ij} - d_e)^2 \\
        \nabla_{\x^i} \gE(d_{ij})  &= 2k_{\alpha} (d_{ij}-d_e) \left(\frac{\y^i-\y^j}{d_{ij}}\right)\\
    \nabla_{\x^j} \gE(d_{ij})  &= -2k_{\alpha} (d_{ij}-d_e)  \left(\frac{\y^i-\y^j}{d_{ij}}\right)\\
    \nabla_{\rvr^i} \gE(d_{ij})  &= 2k_{\alpha} (d_{ij}-d_e) \left(\frac{\y^i-\y^j}{d_{ij}}\right)(\y^{i*})^{\top}\\
    \nabla_{\rvr^j} \gE(d_{ij})  &= -2k_{\alpha} (d_{ij}-d_e)  \left(\frac{\y^i-\y^j}{d_{ij}}\right)(\y^{j*})^{\top}\\
\end{align}
The energy function of the CDRs is
\begin{align}
    \gE_{2} &= \sum_{i}\sum_{j=i+1} \gE(d_{ij})     \\
        \nabla_{\x_i}\gE_{3} &= \sum_{i}\sum_{j=i+1} \nabla_{\x_i}\gE(d_{ij})     \\
    \nabla_{\rvr_i}\gE_{3} &= \sum_{i}\sum_{j=i+1} \nabla_{\rvr_i}\gE(d_{ij})     \\
\end{align}
\paragraph{III) $\text{C}_\alpha^{i}-\text{N}^{i+1}$ energy guidance.}
Here, $d_e\approx 2.42$\char"C5. Let $j=i+1$
\begin{align}
\x^i &=  \x^i\\
\y^j &= \rvr^j \y^{j*} + \x^j;\frac{d\y^j}{d\x^j} = \mbf{1} \\
    d_{ij} &= \|\x^i - \y^j\|_2\\
    \gE(d_{ij}) & = k_{\alpha}(d_{ij} - d_e)^2 \\
        \nabla_{\x^i} \gE(d_{ij})  &= 2k_{\alpha} (d_{ij}-d_e) \left(\frac{\x^i-\x^j}{d_{ij}}\right)\\
    \nabla_{\x^j} \gE(d_{ij})  &= -2k_{\alpha} (d_{ij}-d_e)\left(\frac{\y^i-\y^j}{d_{ij}}\right)\\
    \nabla_{\rvr^j} \gE(d_{ij})  &= -2k_{\alpha} (d_{ij}-d_e)  \left(\frac{\y^i-\y^j}{d_{ij}}\right)(\y^{j*})^{\top}\\
\end{align}
The energy function of the CDRs is
\begin{align}
    \gE_{3} &= \sum_{i}\sum_{j=i+1} \gE(d_{ij})     
\end{align}
\paragraph{IV) $\text{C}^i-\text{C}_\alpha^{i+1}$ energy guidance.}
Here, $d_e\approx 2.44$\char"C5. Let $j=i+1$.
\begin{align}
\y^i &= \rvr^i \y^{i*} + \x; \frac{d\y^i}{d\x^i} = \mbf{1}^i \\
\y^j &=\x^j \\
    d_{ij} &= \|\y^i - \x^j\|_2\\
    \gE(d_{ij}) & = k_{\alpha}(d_{ij} - d_e)^2 \\
        \nabla_{\x^i} \gE(d_{ij})  &= 2k_{\alpha} (d_{ij}-d_e) \rvr^i \left(\frac{\y^i-\y^j}{d_{ij}}\right)\\
    \nabla_{\x^j} \gE(d_{ij})  &= -2k_{\alpha} (d_{ij}-d_e) \left(\frac{\x^i-\x^j}{d_{ij}}\right)\\
    \nabla_{\rvr^i} \gE(d_{ij})  &= 2k_{\alpha} (d_{ij}-d_e) \left(\frac{\y^i-\y^j}{d_{ij}}\right)(\y^{i*})^{\top}\\
\end{align}
The energy function of the CDRs is
\begin{align}
    \gE_{4} = \sum_{i}\sum_{j=i+1} \gE(d_{ij})     
\end{align}

Overall, the \textbf{final energy function} is
\begin{align}
    \gE &= \omega_1 \gE_1 + \omega_2 \gE_2 + \omega_3 \gE_3 + \omega_4 \gE_4 \\
\end{align}
    and the guidance on residue $i$ is:
\begin{align}
    \nabla_{\x_i}  \gE  &= \omega_1 \nabla_{\x_i}\gE_1 + \omega_2 \nabla_{\x_i}\gE_2 + \omega_3\nabla_{\x_i}\gE_3 + \omega_4\nabla_{\x_i}\gE_4 \\
    \nabla_{\rvr_i}  \gE  &= \omega_1\nabla_{\rvr_i}\gE_1 + \omega_2\nabla_{\rvr_i}\gE_2 + \omega_3\nabla_{\rvr_i}\gE_3 + \omega_4\nabla_{\rvr_i}\gE_4 \\
\end{align}
where $\{\omega_i| i=1,2,3,4\}$ are hyper-parameters and set $k_\alpha $ to $1$.
\section{Implementation details and experimental setup}
\label{appsec:experiment_setup}
\subsection{Training details}\label{appsubsec:Hyperparameters}
All models are trained with SGD using the ADAM optimizer.
The training process takes around 72 hours with two Nvidia 3090 GPUs with 24GB RAM.
The range of hyperparameters can be seen in \autoref{tab:hyper-parameter}

\begin{table}[ht] 
\centering
\caption{Hyper-parameter search range for antibody refinement.} 
\vspace{1mm}
\resizebox{0.5 \linewidth}{!}{
\begin{tabular}{l| l|l l}
\toprule
     Methods & Hyper-parameter        & Range           & best    \\ 
\midrule
    \multirow{15}{*}{Training}
& residue feature dimension    & \{128\}  & 128 \\
    & pair feature dimension & \{64\}  & 64 \\
    & IPA Layers L & \{4,6\} &  \\
    & Batch size & \{8,16\} & 8 \\
    \cline{2-4}
    & Optimizer & Adam   & - \\
    & Weight decay & \{1e-4\} & 1e-4\\
    & $\beta_1$ & \{0.9\}  & \\
    & $\beta_2$ & \{0.999\} &  \\
    & Learning rate        & \{0.0001\} &    0.0001              \\
    & Epochs & \{300k, 500k,700k\}  & - \\
    & lr scheduler & plateau \\
    & lr scheduler patience & \{10,20,30\}  & - \\
    & lr scheduler factor & 0.8  & - \\
    & lr scheduler min lr & 5.e-6  & - \\
    & GPU & RTX 3090  & - \\
    & $\lambda$ & \{0, 0.1, 1, 10\}& \\
    \midrule
    \multirow{5}{*}{Sampling}    
    & $g(t)^2$ & \{1\}  \\
    & $\beta$ & \{0.01,0.05,0.1,0.12,0.15\} \\
    & $i(t)$ & \{0.1,0.5,1\}  \\
    & time steps & 2  & -\\    
    & $\omega_i$ & {0.0,0.5,1.0}  & -\\    
\bottomrule                     
\end{tabular}  
}
\label{tab:hyper-parameter}
\end{table}

\subsection{Baselines.} The results of RepertoireBuilder~\cite{ruffolo2023fast}, AlphaFold-Multimer~\cite{evans2021protein}, DeepAb~\cite{ruffolo2022antibody} (w. Rosetta), ABlooper~\cite{abanades2022ablooper} (w. OpenMM),  and IgFold~\cite{ruffolo2023fast} (w. Rosetta) are adopted from the reported values in the IgFold paper.
For AlphaFold3~\cite{abramson2024accurate}, we directly make the structure prediction by the \href{https://alphafoldserver.com/}{Alphafold server}.

\subsection{Why do we need DiffAB split?}
For DiffAB~\citep{luo2022antigen}, it can be regarded as a structure prediction method that only predicts the CDRs. Therefore, we can use it as the prior for flow matching. However, the checkpoint provided in the public GitHub is trained by the DiffAB split. Thus, to fairly verify that the proposed method can improve the quality of structure generated from DiffAB, we involve the DiffAB split in the experiments. Additionally, we also use ABlooper as the prior with DiffAB split to demonstrate the robustness and generalization of our method.

\subsection{SDE training and inference}\label{app:sec:sde_train_inf}
In our experiment, we also consider the stochastic version of $\sethree$ flow matching~\cite{bose2024sestochastic}, which incorporates the stochastic term in the interpolated state $\x_t,\rvr_t$,
\begin{align}
\label{eq:sde_x_t}
    &\tilde{\rvr}_t \sim\igso(\rvr_t, \gamma_\rvr^2(t) t(1-t)) \\
&\tilde{\x}_t \sim \mathcal{N}(\x_t, \gamma_\x^2(t) t(1-t)).
\end{align}
This form is also called the Schrödinger bridge. A more simplified version is $\tilde{\x}_t \sim \mathcal{N}(\x_t,\sigma\mbf{I})$.
We outline our training and inference algorithms for the $\sothree$ component of \flowab. The training algorithm is detailed in Alg. \ref{alg:flowab_training_sde} while the inference algorithm is provided in Alg. \ref{alg:sfm_inference_sde}.

\begin{algorithm}
\caption{\flowab training with stochasticity}
\label{alg:flowab_training_sde}
\begin{algorithmic}[1]
\State \textbf{Input:} Source and target $p_0(\x_0), p_1(\x_1)$, flow network $v_\theta$, and diffusion scalings $\gamma_\rvr(t)$, $\gamma_\x(t)$.
\While{Training}
    \State $t, \x_0, \x_1 \sim \gU(0, 1), p_0, p_1$

    \State $\rvr_t \leftarrow \exp_{\rvr_0}(t \log_{\rvr_0}(\rvr_1))$ \Comment{geodesic interpolant from
    }
    \State $\x_t \leftarrow t \x_0 + (1 - t) \x_1$
    \Comment{interpolant (Euclidean)}
    
    \State $\tilde{\rvr}_t \sim\igso(\rvr_t, \gamma_\rvr^2(t) t(1-t))$ \Comment{simulation-free approximation from \autoref{eq:sde_x_t}
    }
    \State $\tilde{\x}_t \sim \mathcal{N}(\x_t, \gamma_\x^2(t) t(1-t))$
    \State ${\mathcal{L}}_{\flowab} \leftarrow \left \|v_\theta(\tilde{\rvr}_t,\gC, t) - \frac{\log_{\tilde{\rvr}_t}(\rvr_1)}{1-t} \right \|_{\sothree}^2 + \left \| v_\theta(\tilde{\x}_t,\gC, t) - \frac{\x_1-\tilde{\x}_t }{1-t} \right\|^2 $
    \State $\theta \leftarrow \text{Update}(\theta, \nabla_\theta \mathcal{L}_{\flowab,})$
\EndWhile
\State \textbf{return} $v_\theta$
\end{algorithmic}

\end{algorithm}

\begin{algorithm}
\caption{\FlowAB inference with stochasticity}
\label{alg:sfm_inference_sde}
\begin{algorithmic}[1]
\State \textbf{Input:} Source distribution \( \rho_0 \), flow network \( v_\theta \), diffusion schedule \( \gamma(\cdot) \), inference annealing \( i(\cdot) \), noise scale \( \zeta \), integration step size \( \Delta t \), energy guided weight $g(t)$.
\State Sample \( (\x_0,\rvr_0) \sim p_0 \)
\For{ \( s\) in \( (0, 1 / \Delta t] \)}
    \State $t \gets s \Delta t$
    \State Sample \( z \sim \mathcal{N}(0, 1) \), \( dB_t \leftarrow \zeta \gamma_t \cdot \sqrt{dt} \cdot z \)

    \State \( d\hat{B}_t \leftarrow \mathrm{hat}(dB_t)\)
    \Comment {map rotation vector to $\sothreelie$}
    \State \( u_t \leftarrow \rvr_t^\top (v_\theta(t, \rvr_t)- \frac{1}{2} g(t)^2\nabla_{\rvr_t}\gE(\backbone_t)) \)
    \Comment {parallel-transport the vector field to $\sothreelie$}
    \State \( \rvr_{t+\Delta t} \leftarrow \rvr_t \exp(u_t i_t \Delta t + d\hat{B}_t)  \)\Comment {rotation update}
    \State  $\x_{t+\Delta t} \leftarrow  \x_t +( v_\theta(t,\x_t)- \frac{1}{2} g(t)^2\nabla_{\x_t}\gE(\backbone_t))\Delta t$ \Comment {position update}
\EndFor
\State \textbf{return} \( \rvr_1, \x_1 \)
\end{algorithmic}
\end{algorithm}

\subsection{Vector field parameterization}
\label{app:vector_field_parameterization}

In FlowAB, we use the IPA from AF2 as the core module for feature extraction~\citep{anand2022protein}. Since the prior structures usually have a high similarity with the target structures, we consider using the antibody complex feature to predict the relative transformation from the priors (i.e. $\x_0$ and $\rvr_0$ as the standing point) to the target (i.e. ground truth $\x_1$ and $\rvr_1$):     

\begin{align}
\hat{\x_1} &= \rvr_0\operatorname{MLP}_1(\mathbf{h}_i,t) + \x_0 \\
\hat{\rvr}_1 &= \rvr_0\operatorname{MLP}_2(\mathbf{h}_i,t) 
\end{align}
One can also use the $\rvr_t$ and $\x_t$ to be the standing point if the prior is significant distinct with the target, such as Gaussian prior or Harmonic prior.
Then, we parameterize the vector field by directly regressing the target $\x_1$ and $\rvr_1$. Thus, the predicted vector fields $\hat v_{\sR^3}$ and $\hat v_{\sothree}$ for sampling are: 
\begin{align}
    \hat{v}_\sR (\x,t) &= f_{\sR^3} (\mathbf{h}_l,t) = \frac{\hat{\x_1}-\x_t}{1-t}\\
    \hat{v}_\sothree (\rvr,t) &=f_{\sothree} (\mathbf{h}_l,t)  =\frac{\log_{\rvr_t} \hat{\rvr}_1}{1-t}
\end{align}

The losses for training is:
\begin{align}
    \gL_{\rvr} &= ConsineEmbeddingLoss(\rvr_1,\hat\rvr_1) \\
    \gL_{\x} & = \|\x_1-\hat\x_1\|_2    
\end{align}
where $ConsineEmbeddingLoss$ is the Consine Embedding Loss that works better than the L-2 loss. This loss can adopt the function \texttt{F.Consine\_embedding\_loss} implemented in PyTorch. We find it can reduce variance and numerical instability in our experiments. Additionally,
we weight the rotation loss with a coefficient $0.5$ 
as compared to the translation loss that has a weight $1.0$.

\subsection{RMSD}\label{appsubsec:rmsd}
Given two sets of $n$ points $\mathbf {v}$ and 
$\mathbf {w}$, the RMSD (root mean square deviation) is defined as follows:
\begin{align}
\mathrm {RMSD} (\mathbf {v} ,\mathbf {w} )&={\sqrt {{\frac {1}{n}}\sum _{i=1}^{n}\|v_{i}-w_{i}\|^{2}}}\\&={\sqrt {{\frac {1}{n}}\sum _{i=1}^{n}((v_{ix}-w_{ix})^{2}+(v_{iy}-w_{iy})^{2}+(v_{iz}-w_{iz})^{2}}})
\end{align}

In our experiments, we calculate the RMSD with four heavy atoms $[\text{N},\text{C}_{\alpha},\text{C},\text{O}]$ as suggested in IgFold~\cite{ruffolo2023fast}.

\end{document}